\documentclass[letterpaper, 10 pt, conference]{ieeeconf}  

\IEEEoverridecommandlockouts  
\usepackage{amsmath}
\usepackage{amssymb}
\usepackage[nospace]{cite}
\usepackage{url}
\usepackage{xcolor}
\usepackage{fancyhdr}
\usepackage[hidelinks]{hyperref}

\usepackage{adjustbox}
\usepackage{graphicx}
\usepackage{tensor}

\usepackage[utf8]{inputenc}
\usepackage[OT1]{fontenc}
\usepackage[final]{microtype}

\usepackage{booktabs}

\usepackage[binary-units]{siunitx}

\usepackage{lipsum}

\title{\LARGE \bf
Marsupial Walking-and-Flying Robotic Deployment for Collaborative Exploration of Unknown Environments
}

\author{Paolo De Petris$^{1,\star}$, Shehryar Khattak$^{2,\star}$, Mihir Dharmadhikari$^{1}$, Gabriel Waibel$^{2}$, Huan Nguyen$^{1}$\\ Markus Montenegro$^{2}$, Nikhil Khedekar$^{1}$, Kostas Alexis$^{1}$, and Marco Hutter$^{2}$ 
\thanks{
}
\thanks{$^{\star}$ The authors have contributed equally.}
\thanks{$^1$NTNU, O. S. Bragstads Plass 2D, 7034, Trondheim, Norway}
\thanks{$^2$ETH Zurich, Leonhardstrasse 21, 8092, Zurich, Switzerland}%
\thanks{Correspondence email: \tt\small paolo.de.petris@ntnu.no}
}

\begin{document}

\maketitle
\thispagestyle{empty}
\pagestyle{empty}

\begin{abstract}
This work contributes a marsupial robotic system-of-systems involving a legged and an aerial robot capable of collaborative mapping and exploration path planning that exploits the heterogeneous properties of the two systems and the ability to selectively deploy the aerial system from the ground robot. Exploiting the dexterous locomotion capabilities and long endurance of quadruped robots, the marsupial combination can explore within large-scale and confined environments involving rough terrain. However, as certain types of terrain or vertical geometries can render any ground system unable to continue its exploration, the marsupial system can --when needed-- deploy the flying robot which, by exploiting its $3\textrm{D}$ navigation capabilities, can undertake a focused exploration task within its endurance limitations. Focusing on autonomy, the two systems can co-localize and map together by sharing LiDAR-based maps and plan exploration paths individually, while a tailored graph search onboard the legged robot allows it to identify where and when the ferried aerial platform should be deployed. The system is verified within multiple experimental studies demonstrating the expanded exploration capabilities of the marsupial system-of-systems and facilitating the exploration of otherwise individually unreachable areas.

\end{abstract}

\section{Introduction}\label{sec:intro}

\begin{figure}[ht!]
\includegraphics[width=\columnwidth]{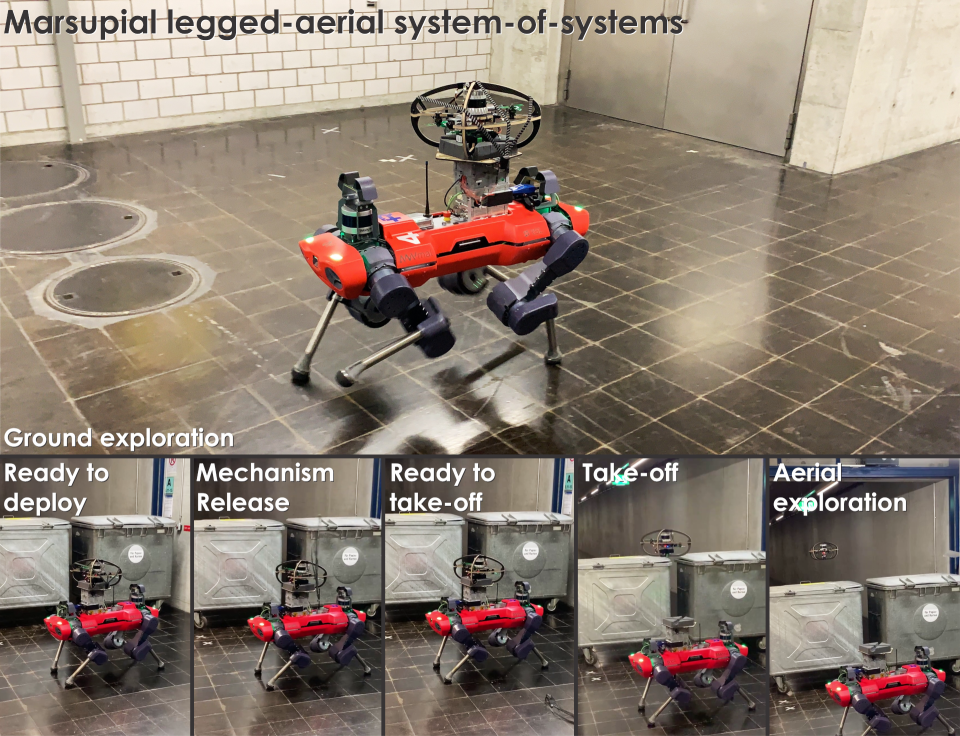}
\vspace{-4ex}
\caption{Instance of the developed marsupial integrated legged-and-aerial robotic system-of-systems tailored for the large-scale autonomous exploration and mapping applications.}
\label{fig:intro}
\end{figure}

Robotic systems have proven their value in exploration and inspection tasks across a variety of environments~\cite{CERBERUS_SUBT_PHASE_I_II,GBPLANNER_JFR_2020,agha2021nebula,englot2013three,murphy2011robot,explorer_phase_i_ii}. Legged and flying robots, in particular, have demonstrated their advanced potential in this domain. Legged systems present dexterous locomotion capabilities thus allowing to overcome challenging terrain, negotiating narrow-access passages, and enabling the traversal of multi-storey facilities~\cite{miki2022learning,agha2021nebula}, while maintaining operational endurance for long periods. However, being a ground platform, they cannot traverse all types of terrain, overcome all types of geometric structures, pass through windows, or acquire a bird's eye views. On the other hand, aerial systems are unbounded from terrain limitations and can seamlessly navigate in $3\textrm{D}$, yet their limited endurance prevents them from undertaking large-scale tasks in a single deployment. Multi-robot teaming, especially by exploiting the heterogeneity of ground and aerial platforms, provides an avenue to benefit from the complementary advantages of both and overcome their individual limitations. However, if deployed separately their platform-specific disadvantages may remain pertinent. An aerial robot may be most useful in an area far from the deployment point which means that its endurance constraints hinder its utility. Yet, that could be where the legged robot mostly benefits from its supporting skills. Motivated by the above, in this work we develop and demonstrate the marsupial combination of a ground and a flying robot within a system-of-systems approach, shown in Figure~\ref{fig:intro}, being capable of performing autonomous exploration and collaborative mapping. Thus leveraging the long endurance and dexterous locomotion of legged systems, and self-deploying an agile aerial robot when needed and where it truly matters. 

The particular design was motivated by the needs of the DARPA Subterranean Challenge but its potential extends across inspection applications. In this context, the ANYmal legged robot~\cite{hutter2016anymal} is considered as the main unit performing autonomous localization and mapping, as well as exploration path planning thus unveiling most of the map but, naturally, it can face traversability limitations and inability to assume viewpoints at heights. Accordingly, when the geometry of the environment necessitates, ANYmal can deploy the RMF-Owl aerial robot~\cite{depetris2022rmfowl} which it ferries onboard. At start-up, RMF-Owl acquires the most current LiDAR-based map of the environment as reconstructed by ANYmal in which it co-localizes to obtain its initial pose estimate in consistent coordinates. Subsequently, the aerial robot performs autonomous exploration targeted towards an area of interest thus complementing the ground system to which it shares the map built online before returning back to the home location. ANYmal can then, in parallel or sequentially, continue its exploration mission. The marsupial combination is thus presenting complementary navigation capabilities and a collective capacity to explore more space, and reach further than the two systems individually or potentially compared to their co-deployment from the same starting point. To demonstrate the value of the proposed system-of-systems we present a set of experiments where complete exploration is not possible by the ground robot alone, or where the combination allows for faster overall coverage. 


\section{Related Work}\label{sec:related}
Marsupial-based robotic systems have been demonstrated to increase the overall capability of the system by exploiting the complementary strength of each robot~\cite{Marques2015survey}. The work in~\cite{Kalaitzakis2021freshwater} presents the use of a marsupial system consisting of an unmanned surface vehicle (USV) and an unmanned aerial vehicle (UAV) in the inspecting of freshwater ecosystems, combining the advantages of the long operation time of the USV and the increased field of view provided by the UAV. Additionally, marsupial systems with aerial and ground robots are utilized in the context of planetary exploration~\cite{Schuster2020space} and disaster response tasks~\cite{Moore2016nested}. Other works in the domain of marsupial robotic systems' deployment aim to determine the deployment and possible retrieval time for the marsupial system. The work in~\cite{ren2018doubleUAV} considers the problem of deploying two flying robots from an unmanned ground vehicle and solves an optimization problem to minimize the time for reaching multiple target points. A different objective is proposed in~\cite{Fargeas2015information}, where a marsupial system consisting of two aircrafts is tasked to gather information about an object of interest, while minimizing the likelihood of their detection by their opponent. In the context of autonomous exploration, the authors in~\cite{lee2021stochastic} propose a Monte Carlo tree search method using the solution to the sequential stochastic assignment problem as a roll out action-selection policy to address the problem of planning deployment times and locations of the carrier robots. The method is verified offline using the recorded data from the Urban Circuit of the DARPA Subterranean Challenge~\cite{chung2021darpa} to select the locations to deploy the passenger aerial robots and to explore the frontiers that cannot be accessed by the ground robots. Another work in~\cite{Stankiewicz2018motion_planning} presents a hierarchical approach where a high-level planner generates a topological multi-graph, encoding the locomotion capabilities of each robot and multiple low-level planners create optimal trajectories considering each robot's dynamics and constraints. In~\cite{Wurm2010temporal_symbolic}, the problem of multi-robot exploration with marsupial robots is cast as a temporal planning problem which is solved by integrating a temporal symbolic planner that plans the deployment and the retrieval actions combined with a traditional path planner.


\section{Marsupial Robotic Exploration}\label{sec:approach}
The proposed solution of marsupial robotic system-of-systems exploration and mapping relies on the synergy of legged and aerial robots that can share maps bidirectionally, co-localize and perform synergistic exploration exploiting their complementary capabilities. 

\begin{figure}[ht!]
\includegraphics[width=\columnwidth]{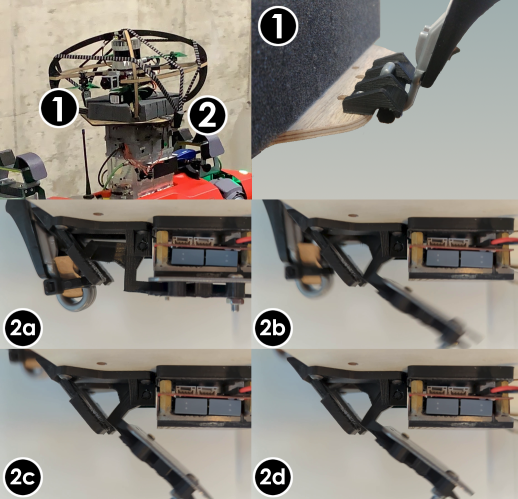}
\caption{Proposed marsupial ground-and-aerial carriage mechanism. Details of the elastic band left side 1) and right side 2) attachment. When the electropermanent magnet is charged 2a), RMF-Owl is firmly secured on the foam base while ANYmal-C continues its exploration path. When the time of deployment is reached the EPM is discharged 2b) and the designed mechanism safely releases the elastic band 2c), 2d).}
\vspace{-2ex}
\label{fig:releasemech}
\end{figure}

\begin{figure*}[ht!]
\includegraphics[width=\textwidth]{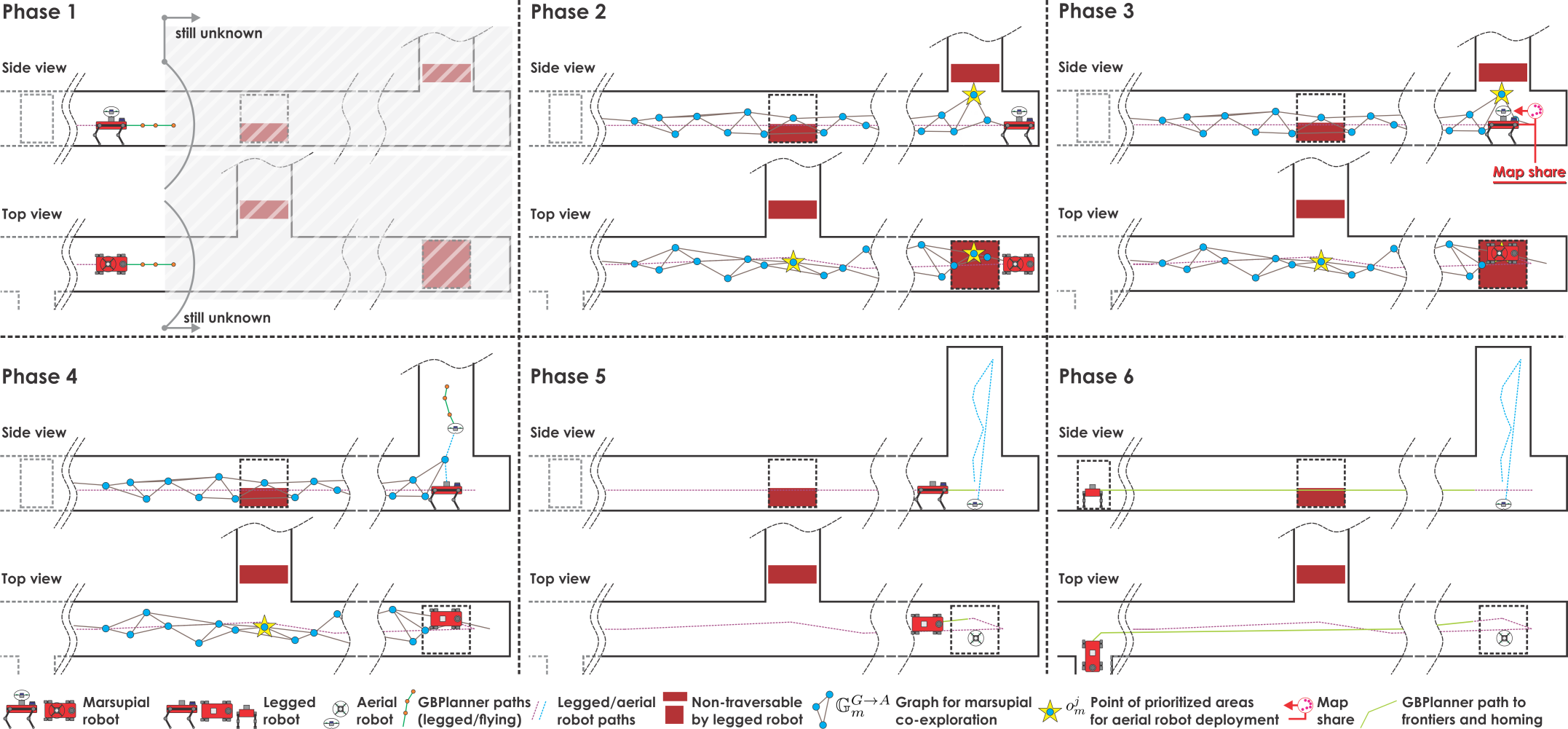}
\caption{Concept visualization of the proposed marsupial ground-and-aerial robotic exploration solution and its key phases. The legged robot initiates exploration, while ferrying the flying system onboard. While individually exploring the environment, it also maintains a sparse graph the vertices of which consider the volumetric gain based on the aerial robot's depth sensor and identifies points where the deployment of the the flying system is beneficial. The legged robot then at a certain point deploys the flying robot. Right before take-off, the two systems share the legged robot's map on which the aerial robot co-localizes and then initiates its individual autonomous exploration focused by the map knowledge of the ground system. Both robots continue exploration individually, share maps also bidirectionally and return to the home location (or other defined setpoint) prior to the depletion of their endurance capabilities.}
\vspace{-2ex}
\label{fig:approach_planner}
\end{figure*}

\subsection{System Overview}
For real-world experiments, the presented approach utilizes a ground-based legged robot, ANYmal-C~\cite{hutter2016anymal}, and an aerial robot, RMF-Owl~\cite{depetris2022rmfowl}, in a marsupial system-of-systems manner, as shown in Figure~\ref{fig:intro}. The ANYmal robot has a payload capacity of \SI{10}{\kilogram} and a continuous operation time of \SI{1}{hour}. For locomotion, the controller proposed by~\cite{miki2022learning} was utilized and ANYmal-C walked with a nominal speed of \SI{0.7}{\metre\per\second} during the experiments. For localization, mapping and path-planning, ANYmal-C utilized measurements from an on-board Velodyne VLP-16 LiDAR, an Epson M-G365PDF0 IMU and kinematic odometry estimates~\cite{tsif}. The VLP-16 sensor offers a Field of View (FOV) of $[360,30]^\circ$ and a maximum range of $100\textrm{m}$. The RMF-Owl, is a lightweight collision-tolerant aerial robot with a custom designed chassis and weighing \SI{1.4}{\kilogram}. This robot has an operation time of \SI{12}{minutes} and during experiments operated with a nominal speed of \SI{1.0}{\metre\per\second}. For all autonomous operations, RMF-Owl utilized an Ouster OS0-64 LiDAR with inertial measurements provided by the flight controller IMU. The OS0-64 offers FOV equal to $[360,90]^\circ$ and a max range of $50\textrm{m}$. The aerial robot is carried on top on the legged robot using a custom design mechanism employing a flexible elastic strap to keep the aerial robot secure during the walking of the legged robot. As depicted in Figure~\ref{fig:releasemech}, the carriage mechanism is equipped with an electropermanent magnet (EPM) permitting the aerial robot to be released when commanded by the planning framework of the legged robot. The proposed design utilizes the tension in the elastic strap to displace it away from the aerial robot upon release to ensure safety during take-off without the need for any additional moving parts and also maintaining a light-weight reusable design. Both robots maintain communication with each other and the base-station using the NimbRo framework~\cite{nimbro}. All map-sharing, co-localization and marsupial exploration planning components were deployed fully on-board each robot and executed in real-time during all missions.

\subsection{Heterogeneous Robot Map Sharing and Co-localization}
To enable collaborative autonomous operation among heterogeneous legged-and-aerial platforms in a marsupial configuration, a bi-directional map sharing and co-localization technique was developed. The proposed method extends our complementary multi-modal localization and mapping framework, CompSLAM~\cite{compSLAM2020}, specifically enabling LiDAR map sharing and co-localization. The key idea behind this extension is that both robots operate in a unified map in a consistent coordinate system, initialized at the deployment position of the legged robot, while maintaining the flexibility for each robot to update parts of the unified map independently during a joint autonomous exploration phase. To facilitate map sharing, similar to map maintenance proposed in~\cite{loam}, the unified map is divided into blocks of volume $10\textrm{m}^3$, with each block identified by a unique hash calculated by the location of each block's center with respect to the origin. Furthermore, each map block is sub-divided into feature sub-blocks containing edge and planar points. During operation, both robots maintain and populate a copy of the unified map onboard, while sharing the same block hash indices among them. Upon request, the map blocks are shared with other robot, updating its on-board map by appending the map blocks with corresponding hash indices.  
To co-localize robots in the unified map, especially during the deployment of the aerial robot from the legged system, the pose of the legged robot in the unified map is also shared with the aerial robot. The shared pose is transformed with respect to the extrinsic mounting orientation of the LiDARs on the legged and aerial robots, and is then used as an initialization guess for the for current pose estimation of the aerial robot in the unified map. The co-localization process utilizes this initialization guess to register the current robot scan with the updated unified map by minimizing point-to-line and point-to-plane correspondences in an iterative manner. Success of the co-localization process is determined by monitoring the incremental pose updates and the remaining number of allowed iterations. If the incremental pose updates become smaller than a pre-defined threshold with the maximum number of allowed iterations not reached, the robot is determined to be co-localized successfully and a success indicator is returned to the state-machine otherwise, in case of non-convergence failure, a warning is raised and the co-localization process has to be restarted.

\subsection{Marsupial Exploration Path Planning}
To enable the autonomous exploration path planning by the marsupial legged-and-aerial robotic system-of-systems, a specialized path planning solution was developed. The method extends our open-sourced Graph-based exploration path planner (GBPlanner)~\cite{GBPLANNER2COHORT_ICRA_2022} which guided all legged and flying robots of Team CERBERUS in the DARPA Subterranean Challenge. GBPlanner allowed the efficient exploration of complex subterranean environments despite the large-scale, at places highly confined and obstacle-filled, multi-level geometries often involved. The method utilizes a volumetric representation of the environment~\cite{voxblox} and employs a bifurcated architecture involving a local and a global planning stage. At the local step, the planner spans at every iteration a dense random graph $\mathbb{G}_l$ within a spatially-defined bounding box and identifies paths that maximize an exploration gain representing primarily the amount of unknown volume to be mapped. However, as geometries can locally be fully mapped (e.g., a dead-end of a mine drift) or because a certain type of robot may be unable to explore all of the local area (e.g., a ground robot will be unable to fly up to a different level through a stope or overcome a tall obstacle), the local stage will --at instances--  report ``completion''. GBPlanner then invokes its global stage which exploits an incrementally built sparse global graph $\mathbb{G}_g$ assembled by selective subsets of the local graphs $\{\mathbb{G}_l\}$. In this sparse global graph, the method can efficiently identify prioritized frontiers, derive optimized paths to such frontiers, and enable auto-homing functionality. 
GBPlanner is used ``as is'' to guide the ground robot in its exploration mission and to perform auto-homing given the robot's VLP-16 range sensor and its FOV $[F_H^G,F_V^G]=[360,30]^\circ$ and a reduced considered maximum range of $d_{\max}^G=20\textrm{m}$, while respecting the traversability limitations of the platform. The two, local and global, graph data structures $\mathbb{G}_l^G,\mathbb{G}_g^G$ are maintained onboard the ground robot. 

The method is however extended in order to guide the marsupial legged-and-aerial robot combination in a manner that exploits their synergies. More specifically, the planner on the ground robot also builds a new sparse global random graph $\mathbb{G}_m^{G\rightarrow A}$ with vertices spanning over the explored $3\textrm{D}$ space. This new graph is distinct than the global graph of GBPlanner responsible for frontier re-positioning and auto-homing. Every vertex $v_m^i$ of $\mathbb{G}_m^{G\rightarrow A}$ is evaluated for exploration gain --being the new previously unmaped volume $\mathcal{V}(v_m^i)$-- observed by that vertex given the aerial robot's depth sensor with FOV $[F_H^A,F_V^A]^\circ = [360,90]^\circ$ and considered effective range $d_{\max}^A=20\textrm{m}$. The vertices $v_m^i$ that are farther than a distance $r_m$ from the current robot location and outside a radius $r_g$ of the frontiers in the ground robot's global graph $\mathbb{G}_g^G$ are clustered and their centers $o_m^j$ are marked as possible regions where the aerial robot can be deployed. 

When GBPlanner on the ground robot reports local completion, the system checks if it has found any areas (and associated points $o_m^j$) where the aerial robot can be deployed. If yes, it moves to the closest region by querying paths to the vertices in $\mathbb{G}_g^G$ closest to these regions, and the associated $o_m^j$, and taking the shortest path among those. Once reached, the ground robot then shares this graph $\mathbb{G}_m^{G\rightarrow A}$ with the aerial robot along with a general exploration bounding box that starts from the current robot height coordinate and expands upwards (or downwards). The GBPlanner on the aerial robot operates similarly to the original method, however every vertex that reaches higher (or lower) in altitude has an increased gain with the goal to promote vertical exploration, while every vertex that spans within the areas spanned in $\mathbb{G}_m^{G\rightarrow A}$ has a decreased gain. Accordingly, GBPlanner on the flying system utilizes the local stage to plan efficiently around the robot by adjusting the gain of the vertices in its graph structure $\mathbb{G}_l^A$, while its global graph $\mathbb{G}_g^A$ is exploited for frontier re-positioning and auto-homing. 

\section{Experimental Results}\label{sec:evaluation}
To systematically evaluate the real-world application and performance of the proposed marsupial system-of-systems approach, experimental tests were conducted in three different environments. The deployment environments emulated real-world challenges such as, exploration of areas unreachable by the ground robot, access to blocked sections of the environment and joint exploration of complex environments with branching paths.

\begin{figure}[ht]
\includegraphics[width=\columnwidth]{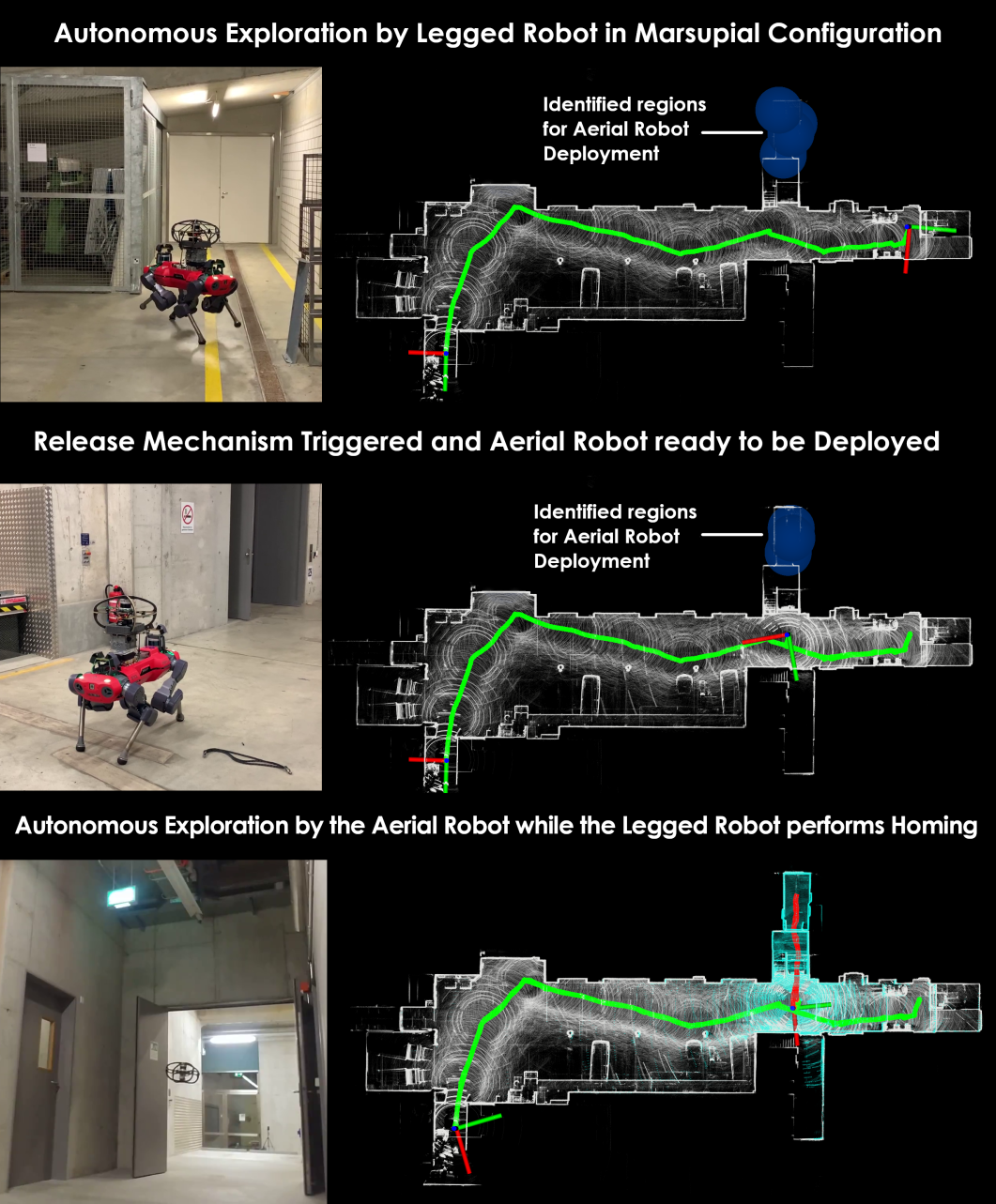}
\caption{Autonomous exploration of an underground garage using heterogeneous robot system-of-systems. Top row shows ANYmal and RMF-Owl starting the mission in marsupial configuration and the identification of the potential regions for the aerial robot deployment. The middle row shows the triggering of the release mechanism at one of the aerial robot deployment location. The bottom row shows both the maps and the exploration paths of ANYmal (Green) and RMF-Owl (Red) in different sections of the environment. The final onboard map of ANYmal contains the region explored by RMF-Owl.}
\vspace{-2ex}
\label{fig:garage}
\end{figure}

\subsection{Complementary Exploration of Inaccessible Areas}
To demonstrate the benefit of utilizing a system-of-systems approach for autonomous exploration, experiments were conducted in environments with some sections accessible to one type of robot but not the other. The complete exploration of these environments required the utilization of the complementary navigation capabilities of legged and aerial robots. First, the physical structure or path blockage rendered parts of the environment un-traversable for the legged robot and only accessible using an aerial robot. Second, the size of the environments imposed an endurance limitation for sole exploration of these environments by using only the aerial robot, hence, necessitating a marsupial deployment.

\subsubsection{Exploration of an Unreachable Elevated Platform}
The first experiment was conducted in an underground garage at ETH Zurich, as shown in Figure~\ref{fig:garage}. This environment consisted of a garage parking area connected to a mezzanine level with a step height of ~\SI{1.0}{\metre}, making it unreachable for the ANYmal robot. During this mission, the robots started autonomous exploration of the garage in marsupial configuration. Once ANYmal completed exploration of the lower level the only exploration frontier left was on the mezzanine level. At this point, the planning stack of ANYmal issued a command to deploy the aerial robot. The RMF-Owl autonomously took-off from the ANYmal and proceeded to explore the mezzanine level, after which it returned to its start position and autonomously landed. The ANYmal robot returned to its mission start position after deploying the aerial robot as there was no remaining information gain available in the environment for exploration purposes.

\subsubsection{Exploration of a Blocked Environment Section}
\begin{figure}[ht]
\includegraphics[width=\columnwidth]{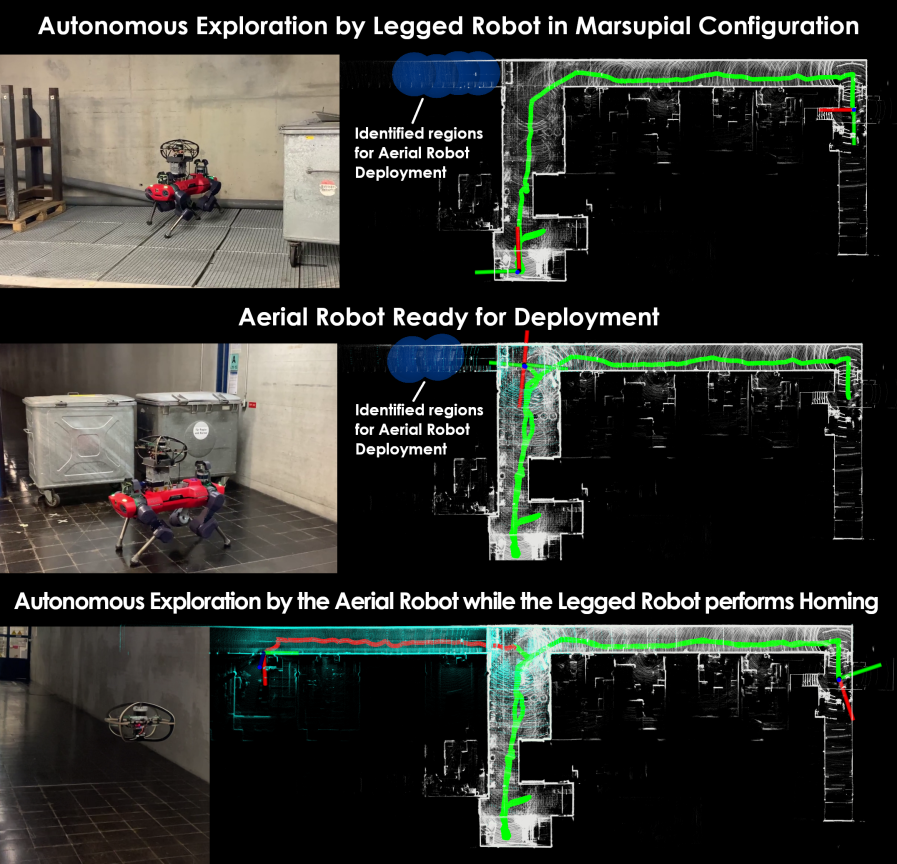}
\caption{Autonomous exploration in marsupial configuration of a branching corridor physically blocked by metal containers. The top row shows the end of the exploration phase of the ANYmal robot while carrying the RMF-Owl. The middle row shows the positioning of the marsupial system at the identified deployment position near the exploration area (blue circles) for the aerial robot. The bottom row shows the autonomous exploration of the aerial robot while the legged robot returns to mission start position.}
\vspace{-4ex}
\label{fig:cla}
\end{figure}

A second experiment was conducted in the basement of the CLA building at ETH Zurich. During this experiment, a branching corridor was physically blocked by metal containers making it inaccessible for the ANYmal robot. However, due to high ceiling clearance, the corridor can be accessed by the aerial robot by flying over the blockage. As shown in Figure~\ref{fig:cla}, the robots start autonomous exploration in marsupial configuration and once all the ground accessible areas are explored by the ANYmal robot, it returns to the site of blockage and chooses to deploy the aerial robot. The aerial robot autonomously takes off and pursues an exploration frontier identified above the blockage. Once RMF-Owl reaches the exploration frontier, it identifies more exploration gain in the corridor behind the blockage and starts autonomous exploration until its endurance limit is reached. After deploying the aerial system, the ANYmal robot returns to mission start point autonomously.

These experiments demonstrate the real-world practical need for a robotic system-of-systems approach for exploration of complex environments by exploiting the complementary navigation capabilities of heterogeneous robots in settings where otherwise complete exploration is not possible.

\begin{figure}[ht]
\includegraphics[width=\columnwidth]{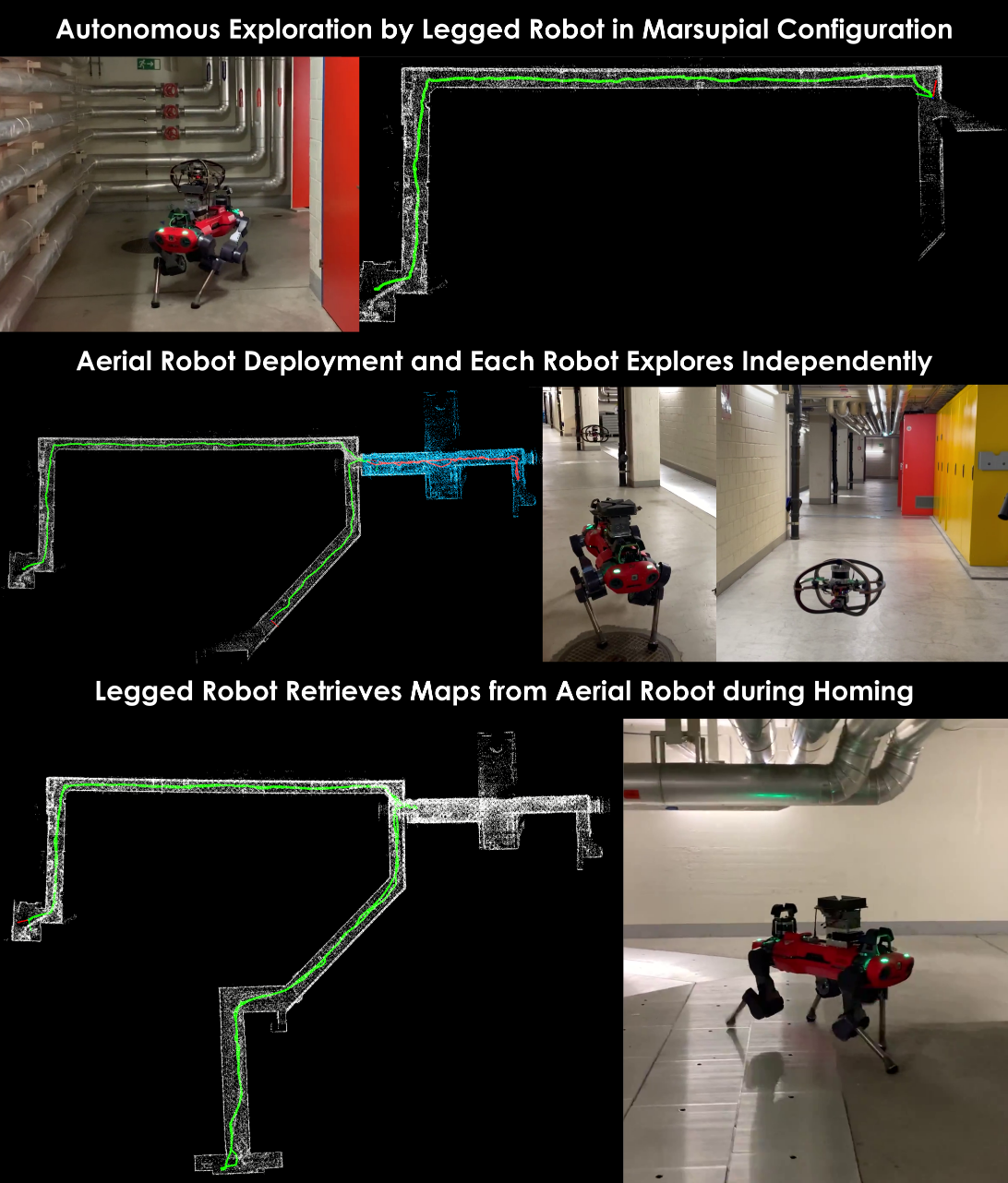}
\caption{Autonomous exploration of multi-branch environments using heterogeneous robots. The top row shows ANYmal and RMF-Owl starting the mission in marsupial configuration and reaching a section of the environment containing diverging paths. The middle row shows deployment of the aerial robot and the map shows the exploration paths of ANYmal (Green) and RMF-Owl (Red) in disjoint branches of the environment. The bottom row shows ANYmal onboard map from ANYmal and its path after autonomously returning to mission start position. The final onboard map of ANYmal contains regions explored by RMF-Owl as they were during homing.}
\vspace{-2ex}
\label{fig:yfloor}
\end{figure}

\subsection{Autonomous Exploration of Multi-branch Environments}
To demonstrate the real-world application potential of marsupial system-of-systems for exploration of multi-branch environments, a test was conducted in the utilities basement of the main building at ETH Zurich. The environment consists of a set of narrow corridors running under the building branching at a certain point to connect the main building basement to that of a neighboring building. The total path length traversed by both robots during autonomous exploration was~\SI{400}{\metre}, with instances of the mission and the full explored map shown in Figure~\ref{fig:yfloor}.
During this experiment, the ANYmal robot in marsupial configuration started autonomous exploration in a small room shown in the top row of Figure~\ref{fig:yfloor} as the leftmost part of the map . After autonomously exploring approximately~\SI{100}{\metre} into the environment the robot detects a branching path, as shown in the top row of Figure~\ref{fig:yfloor}, with potential exploration frontiers detected in diverging directions. At this instance, the legged robot chooses to deploy the aerial robot to pursue exploration along one of the frontiers. Current missions maps and target exploration direction bounds are given from ANYmal to RMF-Owl. After performing co-localization, the legged robot releases the aerial robot's safety strap and RMF-Owl autonomously takes off to start exploration, while ANYmal proceeds to explore the second frontier, as shown by the diverging paths of both robots in the middle row of Figure~\ref{fig:yfloor}. Both robots complete their exploration missions by reaching the dead ends of their individual exploration areas and, as there is no more exploration gain, choose to autonomously return to their initial deployment positions. Please note that the RMF-Owl returns to its known deployment position from the ANYmal robot rather than the start position of the mission as its remaining flight endurance would not have allowed it. While homing towards the mission start position, the ANYmal robot passes by the RMF-Owl and, upon establishing communication, retrieves and appends maps from the aerial robot to its onboard maps, as shown in the in the bottom row of Figure~\ref{fig:yfloor}. This experiment demonstrates the potential and benefit of employing system-of-systems configuration of heterogeneous robots to explore disjointed sections of large-scale and complex environments efficiently.

\section{Conclusions}\label{sec:conclusions}
This work proposed a complete solution for the marsupial legged-and-aerial robotic system-of-systems exploration and mapping deployment. The proposed approach efficiently utilizes the complementary capabilities of walking and flying systems and mitigates their individual limitations by exploiting their synergy. The solution offers bi-directional map sharing, enables co-localization between robots and facilitates collective map building. Simultaneously, the exploration path planning is treated in a unified manner by first allowing each system to explore individually but also enabling the ground robot to decide where and when to deploy the aerial and how to direct its exploration task. Extensive experimental results using the marsupial integration of ANYmal-C and RMF-Owl robots demonstrate that the system-of-systems approach can enable the autonomous exploration and mapping of areas that could not have been fully covered using a single system in an efficient manner.

\addtolength{\textheight}{-12cm}   



\bibliographystyle{IEEEtran}
\bibliography{0_main}

\end{document}